\documentclass{article}

\usepackage{microtype}
\usepackage{graphicx}
\usepackage{subfigure}
\usepackage{booktabs} % for professional tables

\usepackage{multirow}

\usepackage{hyperref}

\usepackage[accepted]{icml2020}

\icmltitlerunning{Soft Labeling Affects Out-of-Distribution Detection of Deep Neural Networks}

\begin{document}

\twocolumn[
\icmltitle{Soft Labeling Affects Out-of-Distribution Detection of Deep Neural Networks}



\icmlsetsymbol{equal}{*}

\begin{icmlauthorlist}
\icmlauthor{Doyup Lee}{postech}
\icmlauthor{Yeongjae Cheon}{kakaobrain}
%\icmlauthor{Wook-Shin Han}{postech}
\end{icmlauthorlist}

\icmlaffiliation{postech}{Pohang University of Science and Technology, South Korea}
\icmlaffiliation{kakaobrain}{Kakao Brain, South Korea}

\icmlcorrespondingauthor{Doyup Lee}{doyup.lee@postech.ac.kr}

\icmlkeywords{Machine Learning, ICML}

\vskip 0.3in
]



\printAffiliationsAndNotice{}  % leave blank if no need to mention equal contribution

\begin{abstract}
Soft labeling becomes a common output regularization for generalization and model compression of deep neural networks.
However, the effect of soft labeling on out-of-distribution (OOD) detection, which is an important topic of machine learning safety, is not explored.
In this study, we show that soft labeling can determine OOD detection performance.
Specifically, how to regularize outputs of incorrect classes by soft labeling can deteriorate or improve OOD detection.
Based on the empirical results, we postulate a future work for OOD-robust DNNs: a proper output regularization by soft labeling can construct OOD-robust DNNs without additional training of OOD samples or modifying the models, while improving classification accuracy.
%DNNS robust  Training robust DNNs to unseen OOD samples is possible by an output regularization only with in-distribution samples and its their soft labeling.

\end{abstract}

\section{Introduction} \label{intro}
%Even though there is no control on test samples after deployment of deep neural networks (DNNs) to the real-world, DNNs often make over-confident predictions for abnormal samples, which are unrecognizable \cite{nguyen2015deep} and from out-of-distribution \cite{hendrycks2017baseline}.
Out-of-distribution (OOD) detection has been an important topic for deep learning applications, after deep neural networks (DNNs) are known to be over-confident on abnormal samples, which are unrecognizable \cite{nguyen2015deep} and from out-of-distribution \cite{hendrycks2017baseline}.
OOD detection is highly related to the safety, because there is no control on test samples after deployment of DNNs to the real-world.

To prevent DNNs from over-confident predictions of OOD samples, post-training with outlier samples is commonly used.
Fine-tuning with a few selected OOD samples \cite{hendrycks2018deep} or adversarial noises \cite{hein2019relu} can improve the detection performance of unseen OOD samples.
%However, as a trade-off between test accuracy and OOD detection, the degradation of classification accuracy is inevitable after the fine-tuning.
%Moreover, we cannot consider all possible OOD samples in training, because there are infinitely many types of OOD.
%In this study, we propose a simple self-distillation framework, improving both OOD detection and classification accuracy. 

Meanwhile, soft labeling becomes a common trick of output regularization to train DNNs in various purposes.
For example, label smoothing \cite{szegedy2016rethinking} improves the test accuracy of DNNs, preventing an overfitting problem \cite{he2019bag, muller2019does}.
Knowledge distillation \cite{hinton2015distilling}, a kind of soft labeling \cite{yuan2019revisit}, can compress the size of a teacher model, or improve the accuracy of its student networks \cite{xie2019self}.

Despite the popularity of soft labeling, how soft labeling affects OOD detection of DNNs has not been explored.
In this study, we assume that regularizing predictions on incorrect classes by soft labeling determines OOD detection performance of DNNs.
We analyze and empirically verify our assumption, based on two major results: a) label smoothing deteriorates OOD detection of DNNs, and b) soft labels, generated by a teacher model, distill OOD detection performance into its student models. %OOD detection performance is distilled into student models by soft labels of their teacher models.
In particular, the degraded test accuracy of a teacher model with outlier exposure is recovered or improved in its student models, while conserving the high performance of OOD detection.

%The teacher model is trained with various OOD samples.
%However, its student model is trained using only ID samples with soft labels, and conserves the outstanding OOD detection performance of the teacher model with outlier exposure.

Based on the empirical results, we claim that a \textit{lottery ticket of soft labeling for OOD-robust DNNs} exists, and how to regularize the predictions of DNNs on incorrect classes is a compelling direction of future work for generalization of DNNs not only on unseen in-distribution (ID) samples, but also on OOD samples.

\section{Preliminaries}
\subsection{Outlier Exposure}
%Exposing a few OOD samples for training significantly improves OOD detection performance \cite{hendrycks2018deep,hein2019relu}.
Outlier exposure \cite{hendrycks2018deep} finetunes a model with some OOD samples to predict uniform distribution for the OOD training samples.
\begin{equation}\label{oe_loss}
    \mathcal{H}(q, p_i) + \lambda \mathcal{H}(\mathcal{U}(K), p_o),
\end{equation}
where $p_i$ is a prediction of a ID sample, $p_o$ is a prediction of a OOD sample, $q$ is an one-hot represented ground truth, $\lambda$ is a hyper-parameter, $\mathcal{H}$ is cross-entropy, and $\mathcal{U}(K)$ is uniform distribution over all $K$ classes.

Despite a significant improvement of OOD detection, training additional OOD samples has two drawbacks. First, original test accuracy is often degraded after outlier exposure as a trade-off between OOD detection and the original task.
Second, we cannot consider all possible OOD samples in training, because there are infinitely many OOD samples.

In this study, soft label prevents the degradation of classification accuracy and often improves the test accuracy (Table~\ref{tab:acc}).
In addition, we show that a soft label can make DNNs robust to OOD without any OOD training sample and model modification (Figure~\ref{OOD_FIGURE}).

%Despite the effectivenss of outlier exposure, there are two limitations in practice.
%First, there are infinitely many OOD samples and we cannot consider all possible OOD samples in training. training.
%Second, outlier exposure degrades original test accuracy instead of the improvement of OOD detection.

\subsection{Soft Labeling as an Output Regularization}
Given an one-hot represented ground truth $q$ of a training sample $x$, soft labeling is defined as
\begin{equation}
    \tilde{q} = (1-\alpha)q + \alpha q',
\end{equation}
where $\alpha$ is a hyper-parameter for soft labeling, $q' \in [0,1]^K$ is a soft target that satisfies $\mathrm{argmax} {(\tilde{q})} = \mathrm{argmax}{(q)}$ and $\sum_{i=1}^{K}{\tilde{q}_i}=1$, and $K$ is the number of classes.
Then, the training loss with soft labeling is
\begin{equation}\label{sl_loss}
    \mathcal{H}(\tilde{q}, p) = (1-\alpha)\mathcal{H}(q, p) + \alpha \mathcal{H}(q', p).
\end{equation}
Note that a soft labeling is a regularization of the predictions including incorrect classes.

The training objective of both label smoothing \cite{szegedy2016rethinking} and knowledge distillation \cite{hinton2015distilling} are represented by Eq~(\ref{sl_loss}) \cite{yuan2019revisit}.
Label smoothing \cite{szegedy2016rethinking} is a soft labeling that regularizes DNNs to predict an uniform distribution $\mathcal{U}(K)$ over all $K$ classes:
\begin{equation}\label{ls_loss}
    (1-\alpha)\mathcal{H}(q, p) + \alpha \mathcal{H}(\mathcal{U}(K), p).
\end{equation}

%As an output regularization, label smoothing makes DNNs predict uniform distribution on incorrect classes:
In knowledge distillation, a soft target of a student model $\tilde{q}$ consists of a prediction of its teacher model:
%When $q'$ is the prediction of a teacher model and $\tilde{q}$ is a training target of its student models, the soft labeling means knowledge distillation.
%In the following sections, we show $\mathcal{H}(q',p)$ can significantly effect on OOD detection of DNNs.
\begin{equation}\label{kd_loss}
    (1-\alpha)\mathcal{H}(q, p) + \alpha \mathcal{H}(p_t, p),
\end{equation}
where $p_t$ is the prediction of the teacher model.
Knowledge distillation is a kind of output regularization of student models by the teacher's predictions \cite{yuan2019revisit} for model compression \cite{hinton2015distilling} or generalization \cite{xie2019self}.

%Label smoothing in the previous section can also be interpreted as a knowledge distillation that a teacher model predicts uniform distribution for all ID samples, $p_t=\mathcal{U}(K)$.

%In terms of OOD detection, we interpret the output regularization of label smoothing as an outlier exposure \cite{hendrycks2018deep} of ID samples.
%Considering ID samples as OOD, label smoothing can deteriorate OOD detection of DNNs if the label smoothing effect only on ID samples, but not on other OOD samples.

\subsection{Experimental Setting}
In this paper, we train WRN-40-2 \cite{zagoruyko2016wide} with the SVHN, CIFAR-10, and CIFAR-100 datasets (ID).
We follow the experimental setting in the official code of outlier exposure\footnote{https://github.com/hendrycks/outlier-exposure} except that we use 150 epochs for training.
In addition, we follow the hyper-parameter settings of knowledge distillation in \cite{muller2019does}.
For evaluation of OOD detection, we use the MNIST, Fashion-MNIST, SVHN (or CIFAR-10), LSUN, and TinyImageNet datasets for OOD samples, and AUROC for the evaluation measure.

\begin{table}
\caption {Test accuracy and expected calibration error (ECE) of WideResNet (Baseline) trained with SVHN, CIFAR10, and CIFAR-100. TinyImageNet is used to train for OE (Outlier Exposure). OD (Outlier Distillation) means the student model of OE model.}
\vskip 0.15in
\centering
\begin{tabular}{lc|ccc}
\toprule
\multicolumn{2}{c|}{ID Dataset}  & Baseline       & +OE    & +OD             \\ \hline
\multirow{2}{*}{SVHN}      & Acc & 97.02          & 96.82 & \textbf{97.17} \\
                           & ECE & 2.38           & 2.65  & \textbf{2.28}  \\ \hline
\multirow{2}{*}{CIFAR-10}  & Acc & \textbf{95.12} & 94.74 & 95.10          \\
                           & ECE & 3.85           & 4.07  & \textbf{3.49}  \\ \hline
\multirow{2}{*}{CIFAR-100} & Acc & 76.63          & 75.58 & \textbf{76.80} \\
                           & ECE & 12.06          & 14.79 & \textbf{10.61} \\ \bottomrule
\end{tabular}
\label{tab:acc}
\end{table}

\section{Soft Labeling Affects OOD Detection} \label{sec:SL_OOD}
\begin{table}[]
\caption {OOD detection performance of outlier exposure and outlier distillation. WRN-OE, which is finetuned with TinyImageNet as OOD, is used as the teacher model for the two student models (WRN and DenseNet).}
\vskip 0.15in
\centering
\begin{tabular}{l|ccc}
\toprule
AUROC         & WRN-OE & $\rightarrow$WRN & $\rightarrow$DenseNet \\ \hline
MNIST         & 84.28   & 90.95             & 93.30                  \\
Fashion-MNIST & 95.16   & 96.03             & 96.61                  \\
SVHN          & 94.19   & 94.50             & 94.19                  \\
LSUN          & 99.99   & 99.94             & 99.94                  \\
TinyImageNet  & 99.99   & 99.78             & 99.83                  \\ \bottomrule
\end{tabular}
\label{tab:dense}
\end{table}

\begin{figure*}[t]
\centering
\includegraphics[width=14.0cm]{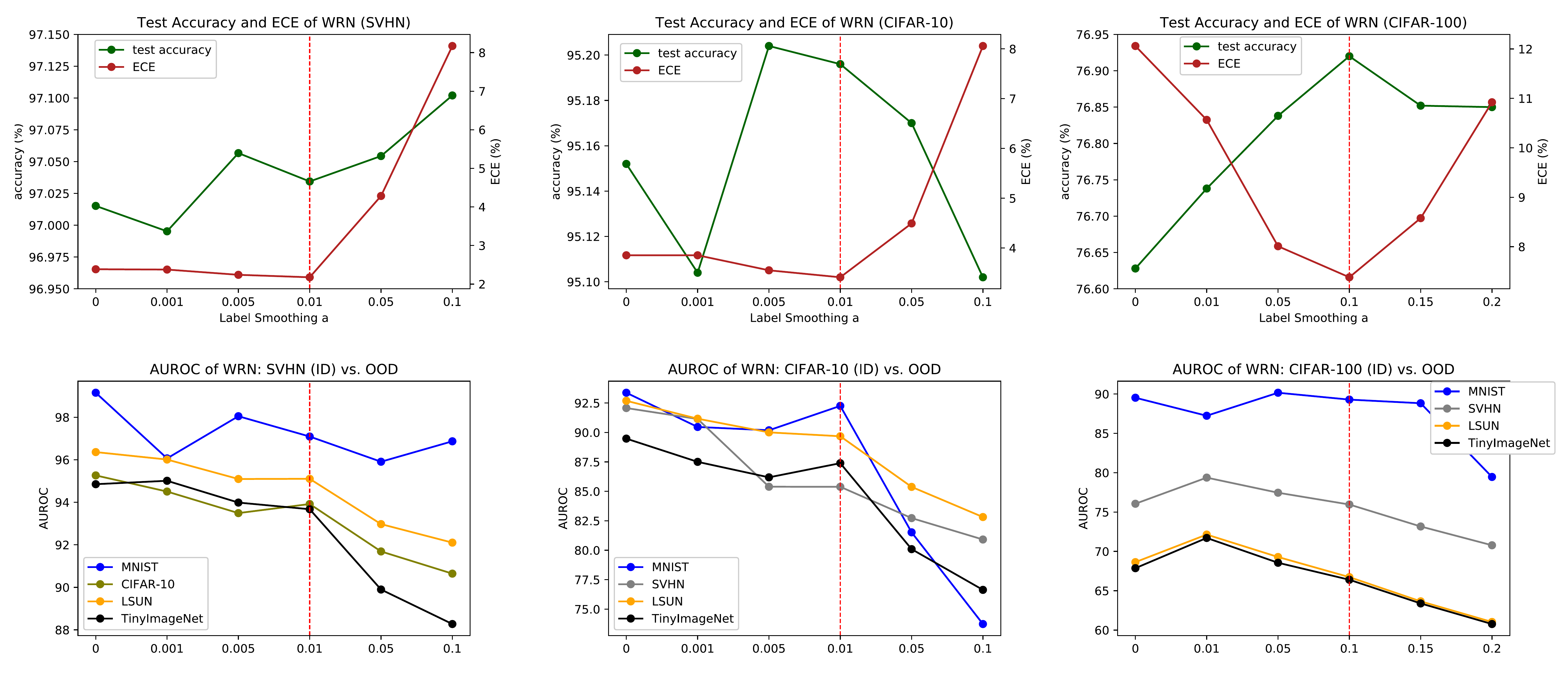}
\caption{Test accuracy and expected calibration error (top) and OOD detection AUROC (bottom) of WRN, trained with SVHN (left), CIFAR-10 (middle), and CIFAR-100 (right) respectively. The red dot line represents label smoothing $\alpha$ minimizing ECE. OOD detection is continuously deteriorated when label smoothing $\alpha$ increases. When ECE starts to increase (after the red dot line), dramatic drops of AUROC are shown in the training datasets of SVHN (ID) and CIFAR-10 (ID).}
\label{LS_FIGURE}
\end{figure*}

\begin{figure*}[t]
\centering
\includegraphics[width=16.0cm]{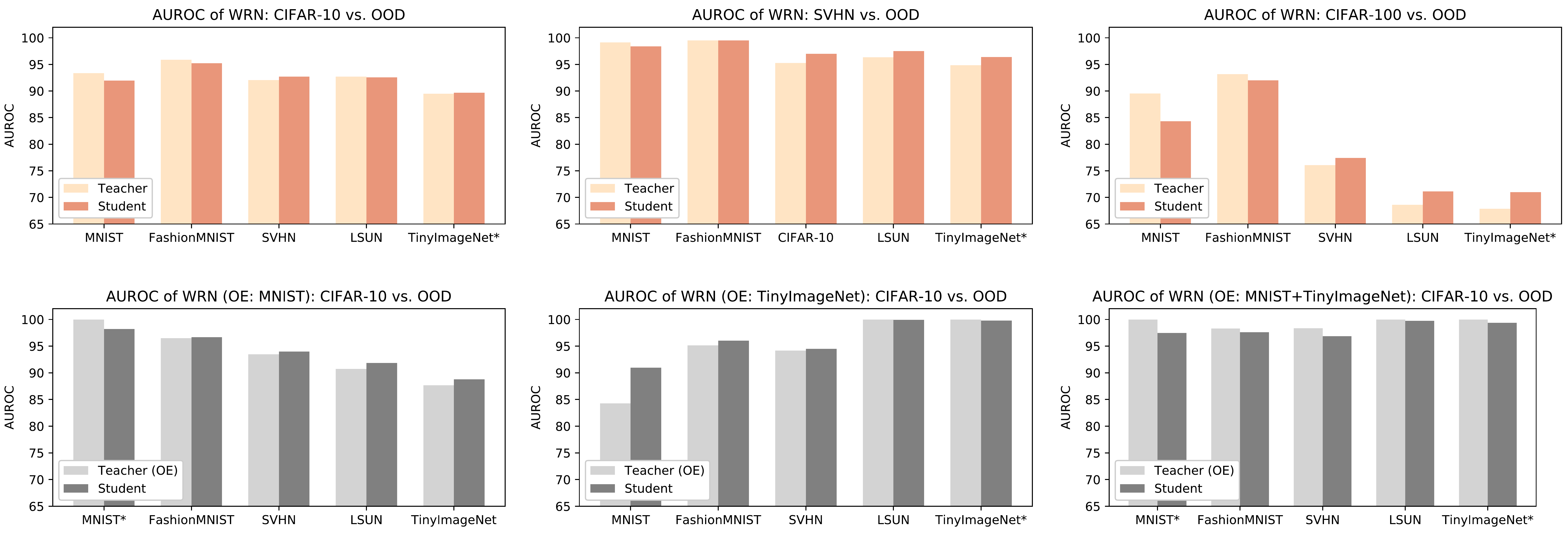}
\caption{OOD detection AUROCs of a teacher model and its student model. (Top) teacher models trained with the SVHN, CIFAR-10, and CIFAR-100 dataset and their student models. (Bottom) teacher models of CIFAR-10 are finetuned with MNIST, TinyImageNet, and MNIST+TinyImageNet by outlier exposure. WRN-40-2 is used as the model architecture of both teacher and student models.}
\label{OOD_FIGURE}
\end{figure*}

\subsection{Label Smoothing and OOD Detection}
Figure~\ref{LS_FIGURE} shows the effects of label smoothing with different $\alpha$ on test accuracy, expected calibration error (ECE) \cite{guo2017calibration}, and detection of the OOD datasets.
As shown in \cite{lukasik2020does}, ECE starts to increase when the label smoothing $\alpha$ is larger than the optimal values (red dot lines).
Although the test accuracy on CIFAR-10 with $\alpha= 0.001, 0.1$ deteriorates, the test accuracy is always improved when ECE is minimized \cite{muller2019does}.

Even though label smoothing improves test accuracy and ECE \cite{muller2019does}, label smoothing makes DNNs vulnerable to out-of-distribution and disable to distinguish ID and OOD datasets.
Label smoothing always deteriorates OOD detection regardless of the magnitude of $\alpha$, and larger $\alpha$ results in more degradation of OOD detection.
In particular, WRN models, trained with SVHN and CIFAR-10, show significant AUROC drops, when the ECE starts to increase (after the red dot line).
%Interpreting the label smoothing as a special case of outlier exposure, the trained models learn to consider ID samples as OOD.

We can infer the reason why label smoothing hurts OOD detection of DNNs from two perspectives.
First, combining Eq~(\ref{oe_loss}) and (\ref{ls_loss}), we can interpret the output regularization of label smoothing as an outlier exposure of ID samples.
Then, label smoothing can deteriorate OOD detection of DNNs, making DNNs disable to discriminate OOD samples from ID samples.
When the magnitude of $\alpha$ increases, the effect of output regularization in Eq~(\ref{ls_loss}) increases and deteriorates the OOD detection performance as outlier exposure of the ID datasets.

Meanwhile, knowledge distillation is the other view to interpret the negative effect of label smoothing on OOD detection.
Note that label smoothing is a knowledge distillation with a teacher model, which perfectly learns the ID samples as OOD and predicts uniform distribution for all ID samples.
Thus, we assume that \textit{Soft labels of incorrect classes, generated by a teacher model, determines OOD detection performance of its student model}, and empirically verify the assumption in section~\ref{sec:KD_OOD}.

\subsection{Knowledge Distillation and OOD Detection} \label{sec:KD_OOD}
In this section, we show that OOD detection performance is determined by the soft labels.
Specifically, soft labels that are generated by a teacher model determine the performance of its student model.
Figure~\ref{OOD_FIGURE} shows OOD detection performance of teacher models and their student models in various settings.
For a student model, we use the same architecture (WRN-40-2) with its teacher, because our concern is to analyze the effects of soft labeling, not a model compression.

Figure~\ref{OOD_FIGURE} (top) shows OOD detection AUROCs of the WRN-40-2 models (SVHN, CIFAR-10, and CIFAR-100), and their student models.
The teacher and its student model have similar AUROCs regardless of test datasets (OOD).

In Figure~\ref{OOD_FIGURE} (bottom), we finetune the teacher models with various OOD samples (MNIST, TinyImageNet, and MNIST+TinyImageNet) to improve OOD detection by outlier exposure.
The OOD detection of the teacher models is improved in different OOD datasets, according to the exposed OOD samples.
We find that OOD detection performance of student models is always consistent with their teacher models (OE), regardless of the choice of training OOD samples for the teacher.
%A student model always has similar OOD detection results to its teacher model, regardless of the choice of training OOD samples for the teacher.

Especially, when we use MNIST+TinyImageNet for outlier exposure of the teacher model, both the teacher and its student almost perfectly detect the test OOD samples.
%Especially, when we use various OOD samples (MNIST+TinyImageNet) to train a teacher model and the OOD detection performance of the teacher is outstanding, its student model has similar and outstanding OOD detection performance.
Exposing various OOD samples in training time is an unrealistic setting, because there are infinitely many cases of OOD.
However, the experimental results is worth noting, because the student model is trained only using ID samples with soft labels, and any OOD sample is not directly used to train the student model.
Although how to generate soft labels without a perfectly OOD-robust teacher remains in an open question, the result show \textbf{the existence of soft labeling for OOD-robust DNNs} to various OOD datasets without OOD training.
%The result imply that there can exist a soft labeling that make DNNs robust to various out-of-distribution samples.

OOD detection performance is also distilled into a student model that has a different architecture from its teacher model.
In Table~\ref{tab:dense}, we use DenseNet \cite{huang2017densely} with 40 hidden layers and 12 growth rates as the student model of WRN-40-2.
Note that the number of trainable parameters of DenseNet (1.1 M) is twice less than WRN-40-2 (2.2 M).
Even though the size and model architecture of the student are different from those of its teacher, OOD detection AUROCs of the teacher and student are consistent.
The results imply that the effect of soft labeling on OOD detection is model-agnostic.
Then, if we find a soft labeling method for OOD-robust DNNs, the soft labeling can be generally used for various DNN architectures.

Orthogonal to the OOD detection, one disadvantage of post-training with OOD samples is a degradation of original classification accuracy \cite{hendrycks2018deep,hein2019relu}.
However, we find that both test accuracy and ECE of the student models (+OD) are similar to or better than the original model before outlier exposure (baseline) in (Table~\ref{tab:acc}).
The improvement of test accuracy results from soft labeling, because soft labels can help model prevent an overfitting problem regardless of the type of soft labeling \cite{yuan2019revisit}.

\section{Discussion}
In this study, we show that a soft labeling of incorrect classes is closely linked with OOD detection of DNNs.
Note that the results of student models in Figure~\ref{OOD_FIGURE} do not use any OOD sample, but can have almost perfect OOD detection AUROCs.
The results verify that constructing OOD-robust DNNs is \textit{possible} without modifying the model or post-training of OOD samples.
%The experimental results in Section~\ref{sec:SL_OOD} support a simple and important assumption: How to make soft labeling of incorrect classes is closely linked with OOD detection of DNNs.
%Note that the student models in Figure~\ref{OOD_FIGURE} do not use any OOD sample, but can have outstanding OOD detection AUROCs.

The limitation of our study is that the solution of soft labeling for OOD-robust DNNs is unrevealed and remains in an open question.
However, we focus on showing the existence of soft labeling for OOD-robust DNNs.

We postulate that \textit{finding an output regularization of incorrect classes that makes DNN robust to unseen OOD samples} is possible and a worth exploration for future work.
Note that proper soft labeling can improve not only OOD detection, but also the classification accuracy of unseen ID samples and confidence calibration (Table~\ref{tab:acc}).
In addition, the OOD-robust soft labeling is model-agnostic and generally applied into various model architectures.

%Although how to generate soft labeling for OOD-robust DNNS without an teacher model is remaining open question, our results show the feasibility of training OOD-robust DNNs only with in-distribution samples.
%When we consider soft labeling can improve test accuracy and ECE, finding a soft labeling method for output regularization that makes DNN robust to unseen OOD samples is worth exploration for future work.

% \usepackage{multirow}

% Note use of \abovespace and \belowspace to get reasonable spacing
% above and below tabular lines.


\nocite{langley00}

\bibliography{example_paper}
\bibliographystyle{icml2020}

\end{document}